\documentclass[sigconf]{acmart}
\usepackage{graphbox}
\usepackage{graphicx, caption, subcaption}

\AtBeginDocument{%
  \providecommand\BibTeX{{%
    \normalfont B\kern-0.5em{\scshape i\kern-0.25em b}\kern-0.8em\TeX}}}

\setcopyright{acmcopyright}
\copyrightyear{2019} 
\acmYear{2019} 
\acmConference[MADiMa '19]{5th International Workshop on Multimedia Assisted Dietary Management}{October 21, 2019}{Nice, France}
\acmBooktitle{5th International Workshop on Multimedia Assisted Dietary Management (MADiMa '19), October 21, 2019, Nice, France}
\acmPrice{15.00}
\acmDOI{10.1145/3347448.3357164}
\acmISBN{978-1-4503-6916-9/19/10}

\acmSubmissionID{8}

\begin{document}
\fancyhead{}
\title{Deep Cooking: Predicting Relative Food Ingredient Amounts from Images}


\author{Jiatong Li}
\email{jiatong.li@rutgers.edu}
\affiliation{%
  \institution{Rutgers University}
  \city{Piscataway}
  \state{NJ}
  \country{USA}
}

\author{Ricardo Guerrero}
\email{r.guerrero@samsung.com}
\affiliation{%
  \institution{Samsung AI Center}
  \city{Cambridge}
  \country{UK}
}

\author{Vladimir Pavlovic}
\email{vladimir@cs.rutgers.edu}
\affiliation{%
  \institution{Rutgers University}
  \city{Piscataway}
  \state{NJ}
  \country{USA}
}

\begin{abstract}
  In this paper, we study the novel problem of not only predicting ingredients from a food image, but also predicting the relative amounts of the detected ingredients. We propose two prediction-based models using deep learning that output sparse and dense predictions, coupled with important semi-automatic multi-database integrative data pre-processing, to solve the problem. Experiments on a dataset of recipes collected from the Internet show the models generate encouraging experimental results.
\end{abstract}




\maketitle

\section{Introduction}
Increased awareness of the impact of food consumption on health and lifestyle today has given rise to novel data-driven food analysis systems, e.g.,~\cite{beijbom2015menu,meyers2015im2calories}, whose goals are to alleviate the challenge of tracking the daily food intake. Many of those systems use data modalities such as images to seamlessly extract information related to the food item that was consumed, often the identity of the meal or its ingredients, or even its caloric value.  While these systems frequently claim to predict the energy intake, they base these predictions on standard energy tables of standardized ingredients (e.g., USDA\footnote{https://ndb.nal.usda.gov/ndb/search/list}). The estimation of the food amount, a highly challenging and often ambiguous task, is delegated to the users themselves. Even the systems that aim to predict a fine-grained ingredient-based representation of the food item, e.g., \cite{salvador2017learning} and \cite{chen2016deep}, do not consider the problem of predicting the ingredients' amounts or relative contributions of the food items in each dish. However, these amounts are paramount for estimating the correct energy value of the meal. A small amount of high-fat food might not be a major health risk factor, while an unhealthy ingredient with a dominant amount may lead to potential health problems.

In this paper, we study the novel problem of predicting the relative amount of each ingredient in a food item from images. To our knowledge, this is the first work goes in to the detail of analyzing the amounts of each ingredients. The problem is both interesting and challenging. It's interesting because we can analyze the nutrients of a food in detail and how each ingredient in the food contributes to health. Some ingredients can be replaced with their low-calorie counterparts and some can be replaced with vegan substitutions according to users' dietary choices. We can also modify the amounts to create healthier foods. It's challenging because the shape and color of even the same ingredient can exhibit large visual differences due to diverse ways of cutting and cooking \cite{chen2016deep}.  Analyzing the amounts of ingredients may also suffer from occlusion. We attack the problem by deep learning models which output the amount of each ingredient.

The contributions of this paper are:
\begin{itemize}
    \item We propose a novel and challenging problem: analyzing the relative amounts of each ingredient from a food image.
    \item We propose prediction-based models using deep learning to solve the problem.
\end{itemize}

\section{Related Work}
As food plays an essential part in our life, there has been a lot of research focusing on food classification, cross-modal retrieval, ingredient analysis and volume estimation.

\noindent{\bf Food Classification.} For classification, initially hand-crafted features and traditional classifiers are used to classify food images. \cite{bossard2014food} uses random forests and SVMs. \cite{beijbom2015menu} uses bag of visual words approach, SVM as the classifier to classify restaurant-specific food and calculate the calories of a meal from a given restaurant.
The traditional methods are outperformed by methods using deep learning features or directly using deep learning.  \cite{ciocca2017food} uses SVM classifier with CNN features. \cite{wang2015recipe} also uses deep learning features for classification. \cite{singla2016food}, \cite{chen2017chinesefoodnet} and \cite{mezgec2017nutrinet} train deep-learning models for food recognition.

\noindent{\bf Cross-Modal Retrieval} is given an image, retrieving its recipe from a collection of test recipes or the other direction: retrieving the corresponding image given a recipe. \cite{salvador2017learning} finds a joint embedding of recipes and images for the image-recipe retrieval task. \cite{marin2019recipe1m+} expands the dataset Recipe1M\cite{salvador2017learning} from 800k images to 13M. However, they expand with images from web search using recipe titles as queries. It is likely that the web images no longer share the same ingredients or instructions as the original ones. They also annotate amount or nutrition information for 50k recipes, while we annotate 250k recipes with our method and 80k of them have images. \cite{chen2018deep} uses attention for finding the embedding for cross-modal retrieval. \cite{wang2019learning} proposes ACME which learns cross-modal embeddings by imposing modality alignment using an adversarial learning strategy and imposing cross-modal translation consistency. 

\noindent{\bf Ingredient Analysis} is predicting the existence of ingredients given a food image. \cite{chen2016deep} uses multi-task deep learning and a graph modeling ingredient co-occurrences. \cite{salvador2019inverse} predicts ingredients as sets and generates cooking instructions by attending to both image and its inferred ingredients simultaneously.
Unfortunately, previous work discard quantity information, losing a lot of important information.

\noindent{\bf Volume Estimation.} Estimating food amounts itself, especially estimating absolute amounts, is not a novel problem. Some methods focus on multi-view 3d reconstruction while others focus on single-view volume estimation.
Multi-view reconstruction can date back to 1994 \cite{shashua1994relative}. Even a more recent work \cite{dehais2017two} using two views uses only traditional computer vision techniques. \cite{zheng2018multi} uses a simple method, mapping contour, but the performance might come from an easy dataset. Although \cite{liang2017computer} uses deep learning in their work, it's only for object detection and they use a simple method to calculate volume.

As for single view methods, \cite{meyers2015im2calories} tries to recognize the contents of a meal from a single image then predicts the calories for home-cooked foods. Their method is based on deep learning for classification and depth estimation to calculate volume and calories. \cite{fang2018single} uses GANs but it requires densely annotated datasets. \cite{ege2019image} reviews three of their existing systems and proposes two novel systems, either using size-known reference objects or foods, including rice grains, or using special mobile devices, like built-in inertial sensors and stereo cameras.

Estimating relative amounts, on the other hand, does not require reference objects or special devices. Our methods estimating relative ingredient amounts can be applied to food images on the internet, where camera information and size-known reference objects are not available.

Furthermore, previous work calculate the amounts of foods as a whole while in our work, we go into the detail and estimate the amount of each ingredient using deep learning models.

\section{Methods}
\label{sec:method}

Let the amounts of ingredients of a recipe be ${\mathbf v_y}=(v_{y_1},\ldots,v_{y_i},\ldots,v_{y_I})$ where $I$ is the total number of ingredients and $v_{y_i}$ is the amount of the $i-th$ ingredient in grams, $v_{y_i}\geq 0.$ $v_{y_i}=0$ when the $i-th$ ingredient is not present in the recipe. Suppose $a$ is a constant and $a>0,~a{\mathbf v_y}$ also represents the same recipe. Therefore we assume that ${\mathbf v_y}$ is normalized such that $\sum_{i=1}^Iv_{y_i}=C$ where $C$ is a constant. $v_{y_i}$ can then be interpreted as the proportion of the $i-th$ ingredient, or there are $v_{y_i}$ grams of the $i-th$ ingredient every $C$ grams of total ingredients. 

Let the amount prediction of an image ${\mathbf x}$ corresponding to the recipe be ${\mathbf v_x}=(v_{x_1},\ldots,v_{x_I})$ and $v_x$ is normalized such that $\sum_{i=1}^Iv_{x_i}=C$. 

We propose two methods, one outputs dense amount predictions and users can threshold according to their applications; the other outputs sparse amount predictions, just like real recipes which usually use a few ingredients. 
\subsection{Dense Method}
When $C=1$, both $v_x, v_y$ can be viewed as probability distributions. If the last layer of a neural network is activated by softmax, the prediction is naturally normalized to $C=1$. Therefore we use softmax to activate the last fully connected layer of a neural network which predicts $v_x$ given ${\mathbf x}$. The loss function to minimize is the cross entropy of $v_x$ and $v_y, L=-\sum_{i=1}^Iv_{y_i}\log v_{x_i}.$
\subsection{Sparse Method}
ReLU activation brings negative values to 0, which brings some sparsity in the output and L1 loss also enforces sparsity. Here, the last fully connected layer of the neural network is activated by ReLU and the loss function is the L1 distance between $v_x,v_y,  L=\sum_{i=1}^I|v_{y_i}-v_{x_i}|$.

\section{Experiments}
\subsection{Dataset}
We use the data in Recipe1M \cite{salvador2017learning} dataset. As the fraction slash is not an ASCII character on recipe websites, it is missing in the dataset after preprocessing. We re-scrape recipe ingredients from websites that are nicely structured, with the quantity and ingredient names parsed.

\noindent{\bf Unit Extraction} We define 3 types of units:
\begin{itemize}
    \item Basic units, including volume, weight and counting units (e.g. box). We define 52 of them and list them in Appendix \ref{append:unit list}.
    \item Size modified units. Usually a counting unit after large/ big/ medium/ small (e.g. small scoop of ice cream). We treat "large/ big" as a multiplier 1.2, "medium" as 1.0 and "small" as 0.8.
    \item Numbers and basic units combined in a parenthesis. (e.g. 1 (15 1/4 ounce) box of cake mix)
\end{itemize}
We only keep recipes that for all ingredients in the recipe, the words between quantity and ingredient name belong to the units we defined. If an ingredient in a recipe does not have quantity field (e.g. some salt), we ignore the ingredient in the recipe. This step only removes 6\% of re-scraped recipes.

\noindent{\bf Canonical Ingredient Construction.} Recipe1M contains about 16k unique ingredients and the top 4k ingredients account for an average coverage of 95\%, which is calculated by $\frac{1}{R}\sum_{k=1}^{R}\frac{i_{ck}}{i_k}$, where $R$ is the number of recipes, $i_k$ is the number of ingredients in the $k-th$ recipe and $i_{ck}$ is the number of ingredients both in the $k-th$ recipe and the top 4k ingredients. The 4k ingredients are further reduced to 1.4k by first merging the ingredients with the same name after a stemming operation and semi-automatically fusing other ingredients. Ingredients are fused if they:
\begin{itemize}
    \item Are close together in Word2vec \cite{mikolov2013distributed} embedding space, which is trained on the titles, ingredients and instructions of Recipe1M.
    \item Share two or three words.
    \item Are mapped to the same item in Nutritionix\footnote{https://www.nutritionix.com/}
\end{itemize}
A human annotator accepts the proposed merger. We only keep recipes with at least 80\% coverage. After the two steps, our data contains 460k recipes in total.

\noindent{\bf Unit Conversion.} The amount numbers are meaningless without converting them to the same unit. In the dataset, one ingredient can be represented by volume, weight and counting units. For example, 2 onions, 1 cup chopped onion, 1 lb onions. We convert the units to grams using Nutritionix and USDA Food Database. The details of mapping can be found in appendix \ref{append:mapping}. Some unit conversions like "1 packet of sugar" are not defined by neither of websites and we only keep recipes where 80\% of the ingredients and their corresponding units are converted. Next we vectorize each recipe into two vectors, $v_y$ for the amount value, described in section \ref{sec:method} and $v_r=(v_{r_1},\ldots,v_{r_I})$ for the amount range, if the amounts are not exactly given. $v_{r_i}$ corresponds to the amount range of the $i-th$ ingredient. For "24-25 ounce cheese\_ravioli", the value is 24.5 ounce, or 686 grams, the range is 28 grams (1 ounce).  Our data contains 250k recipes after unit conversion and 80k of them have images.

\subsection{Compared Methods}
{\bf Retrieval-Based Method:} One method of predicting the ingredients given a food image is cross-modal recipe retrieval which outputs the ingredients and the corresponding amounts of the retrieved recipe. We use the model in \cite{chen2018deep}. For a fair comparison, the model is trained only with ingredients and images. Titles and instructions are not included as the prediction-based models do not use titles or instructions during training. No amount information is used during training the retrieval model. Following \cite{salvador2017learning}, the top 1 recipe among 1000 randomly selected recipes is retrieved. We try two settings, one includes the ground truth recipe in the 1000 recipes and the other without. The two settings only differ by 1 recipe.

\noindent{\bf Prediction-Based Methods:} We use a Resnet50 \cite{he2016deep} pre-trained on UPMC \cite{wang2015recipe} and replace the last layer with ingredient amount prediction. {\bf Softmax:} The dense method in Section \ref{sec:method}. As recipe ingredients are sparse, the dense output is thresholded to top 10 predictions and renormalized. {\bf L1:} The sparse method in Section \ref{sec:method}. Both models are fine tuned with Adam optimizer with learning rate $10^{-4}$. The batch size is 64.

\subsection{Evaluation and Results}
We report 4 evaluation metrics, the first two evaluate ingredient detection and the last two evaluate amounts and calories. If the predicted amount is non-zero for an ingredient, the ingredient is viewed as detected. The ingredient is viewed as not exist if the predicted amount is 0.
\begin{itemize}
    \item {\bf Recall of ground truth ingredients} \# common ingredients between ground truth and predicted over \# ground truth.
    \item {\bf IOU} \# common over \# the union of ground truth and predicted.
    \item {\bf L1 Error} First normalize both the ground truth range vector and the predicted vector to $C=1000$ (every kilogram of total ingredients). The ground truth range vector is normalized according to the ground truth amount vector. If the predicted amount falls outside the range, the error in the dimension is the difference between the amounts. The error is zero otherwise. The $L1$ norm of the error is reported.
    \item {\bf Relative Calorie Error (RCE)} First normalize both the ground truth vector and the predicted vector. Suppose the energy of the i-th ingredient is $c_i~kCal/g$, the ground truth calorie can be estimated as $C_y=\sum_{i=1}^Ic_iv_{y_i}$ and the predicted calorie is $C_x=\sum_{i=1}^Ic_iv_{x_i},$ the error $\frac{|C_y-C_x|}{C_y}$ is reported.
 \end{itemize}
The results are shown in Table \ref{tab:results}. The 80k recipes with images are randomly split into training, validation and test sets, with 48k, 16k, 16k each. The results are on test set. Numbers are "mean (standard deviation over test set)". The up arrows indicate the higher the better and the down arrows indicate the lower the better.
\begin{table}[htp!]
    \centering
    \begin{tabular}{|p{1.3cm}p{1.4cm}p{1.4cm}p{1.3cm}p{1.2cm}|}
    \hline
         & \textbf{Recall$\uparrow$}&\textbf{IOU$\uparrow$}&\textbf{L1$\downarrow$}&\textbf{RCE$\downarrow$}\\
         \hline
         Retrieval w/o gt & 0.26 (0.23) & 0.16 (0.16)& 1613.65 (469.17)& 1.00 (38.47)\\
        Retrieval w/ gt & 0.33 (0.31) & {\bf 0.24 (0.29)}& 1468.48 (653.69)& 0.94 (38.47)\\
        L1 & 0.32 (0.19) & 0.21 (0.15) & 1498.05 (476.83)&0.87 (8.07)\\
        Softmax & {\bf 0.35 (0.31)} & 0.17 (0.11) & {\bf 1433.61 (444.22)} & {\bf 0.74 (10.53)}\\
        \hline
    \end{tabular}
    \caption{Results of the methods}
    \label{tab:results}
\end{table}

The results show that there is no significant difference between the performance of prediction based methods and retrieval with the ground truth recipe included while they all outperform retrieval without ground truth in terms of recall and L1 loss. Compared with retrieval with ground truth, prediction-based methods have lower standard deviations in IOU, L1 and RCE, or more robust performances in general. This is reasonable as retrieval based methods can be affected by the collection of recipes available for the retrieval system. 

\begin{figure}
    \centering
    \begin{minipage}[t]{.18\linewidth}
    \includegraphics[align=t,width=1.0\textwidth]{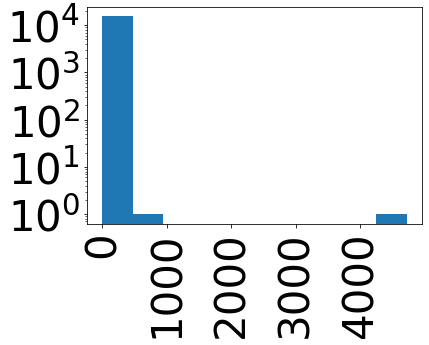}
    \end{minipage}
    \begin{minipage}[t]{.18\linewidth}
    \centering
        \includegraphics[align=t,width=1.0\textwidth]{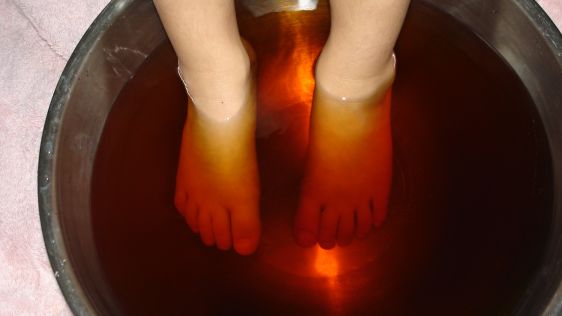}
    \subcaption*{\centering Coffee foot soak}
    \end{minipage}
    \begin{minipage}[t]{.18\linewidth}
    \centering
    \includegraphics[align=t,width=1.0\textwidth]{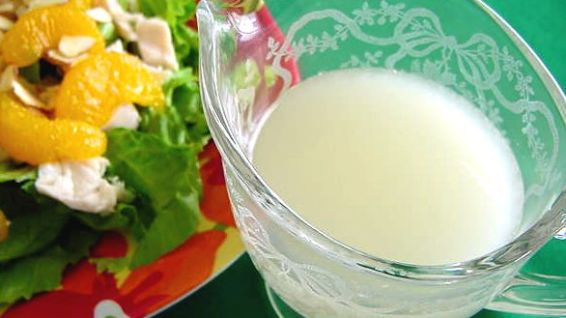}
    \subcaption*{\centering Almond Salad dressing}
    \end{minipage}
    \begin{minipage}[t]{.18\linewidth}
    \centering
    \includegraphics[align=t,width=1.0\textwidth]{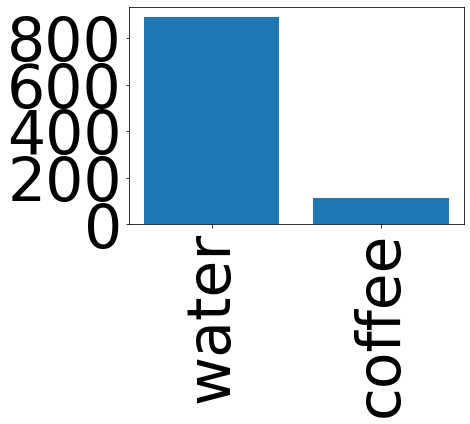}
    \end{minipage}
    \begin{minipage}[t]{.18\linewidth}
    \centering
    \includegraphics[align=t,width=1.0\textwidth]{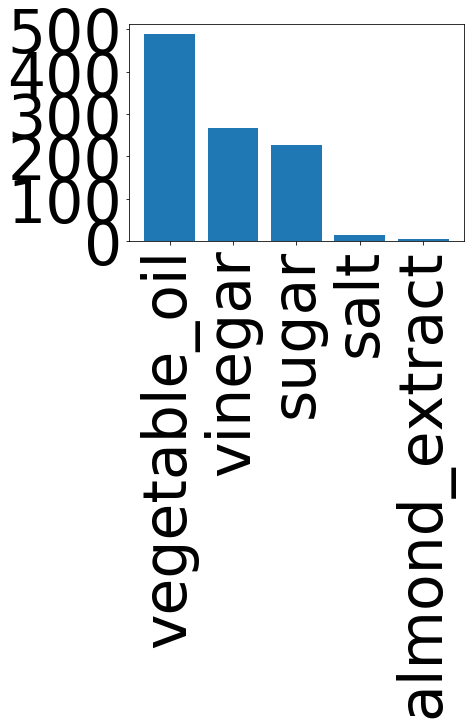}
    \end{minipage}
    \caption{Analysis of relative calorie error. Left to right: histogram of relative calorie error, ground truth image of the recipe with the largest RCE, retrieved image, ground truth amounts, retrieved amounts. The amounts are normalized to $C=1000$.}
    \label{fig:maxrce}
\end{figure}

Relative calorie error tend to have a larger standard deviation because low calorie recipes are sensitive to calorie differences. Figure \ref{fig:maxrce} shows the histogram of relative calorie errors in log scale and the recipe with the largest relative calorie error 4725.32. Most recipes have small relative calorie errors while there are some outliers. The error of about 88\% of the test recipes is less than 1. The calorie of the recipe with the largest RCE after normalizing the sum of ingredients is about $1.11 kCal$ and the calorie of the retrieved recipe is about $5255.65 kCal$. The system retrieves the salad dressing recipe probably because both recipes are liquid.

When thresholded to top 10 predictions and renormalized, the Softmax method performs well in terms of amounts and calories. Furthermore, the threshold can be adjusted according to users' needs, showing the method is able to produce encouraging results at the same time being flexible.

\begin{figure*}[t!]
    \centering
    \begin{minipage}[t]{.15\linewidth}
    \centering
    \includegraphics[align=t,width=1.0\textwidth]{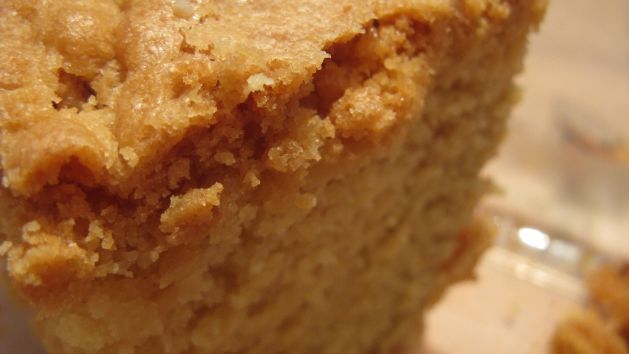}
    \subcaption*{\centering Original pound cake}
    \end{minipage}
    \begin{minipage}[t]{.15\linewidth}
    \centering
    \includegraphics[align=t,width=1.0\textwidth]{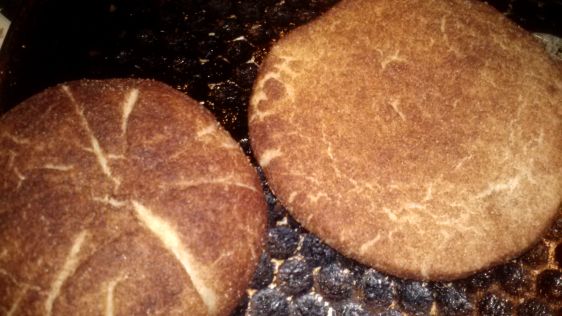}
    \subcaption*{\centering Super soft snickerdoodle cookies}
    \end{minipage}
    \begin{minipage}[t]{.15\linewidth}
    \centering
    \includegraphics[align=t,width=1.0\textwidth]{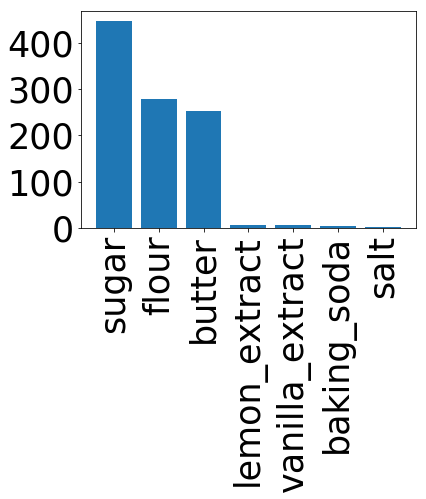}
    \end{minipage}
    \begin{minipage}[t]{.15\linewidth}
    \centering
    \includegraphics[align=t,width=1.0\textwidth]{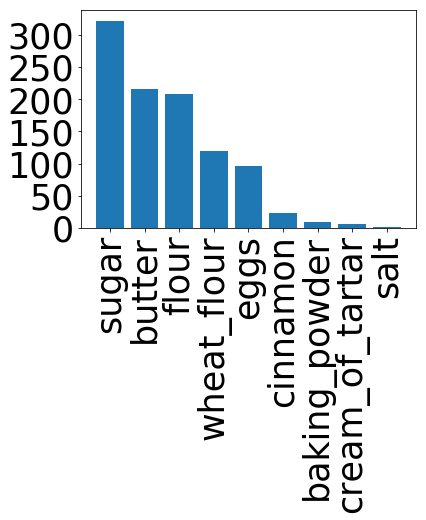}
    \end{minipage}
    \begin{minipage}[t]{.15\linewidth}
    \centering
    \includegraphics[align=t,width=1.0\textwidth]{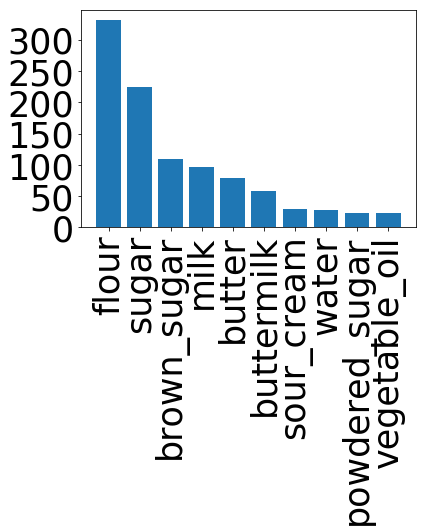}
    \end{minipage}
    \begin{minipage}[t]{.15\linewidth}
    \centering
    \includegraphics[align=t,width=1.0\textwidth]{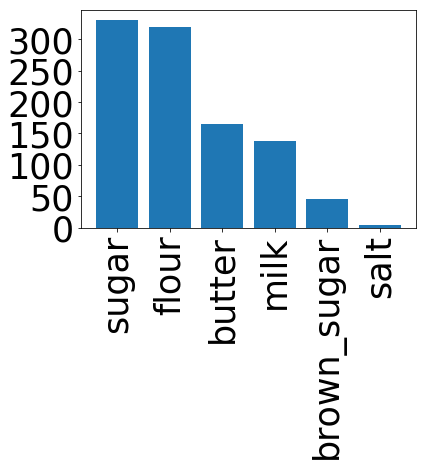}
    \end{minipage}
    \begin{minipage}[t]{.15\linewidth}
    \centering
    \includegraphics[align=t,width=1.0\textwidth]{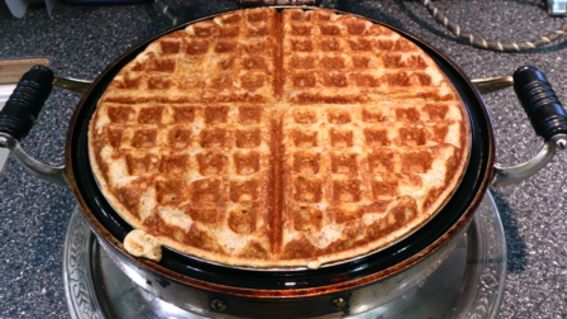}
    \subcaption*{\centering Buttermilk cornmeal waffles}
    \end{minipage}
    \begin{minipage}[t]{.15\linewidth}
    \centering
    \includegraphics[align=t,width=1.0\textwidth]{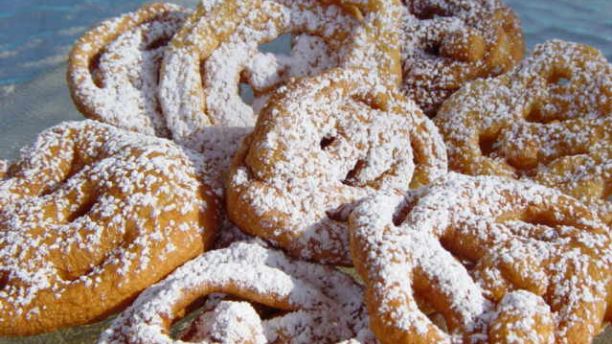}
    \subcaption*{\centering Tippaleiv\"{a}t-May Day fritters}
    \end{minipage}
    \begin{minipage}[t]{.15\linewidth}
    \centering
    \includegraphics[align=t,width=1.0\textwidth]{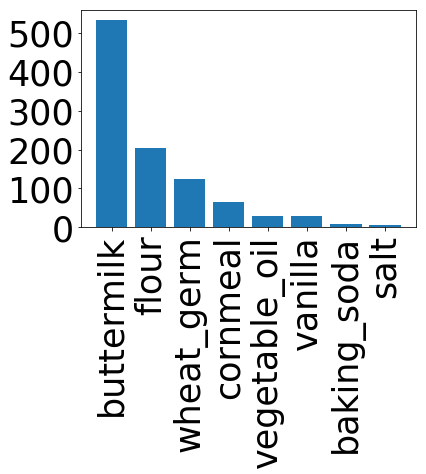}
    \end{minipage}
    \begin{minipage}[t]{.15\linewidth}
    \centering
    \includegraphics[align=t,width=1.0\textwidth]{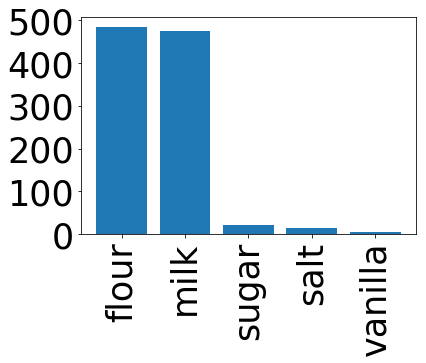}
    \end{minipage}
    \begin{minipage}[t]{.15\linewidth}
    \centering
    \includegraphics[align=t,width=1.0\textwidth]{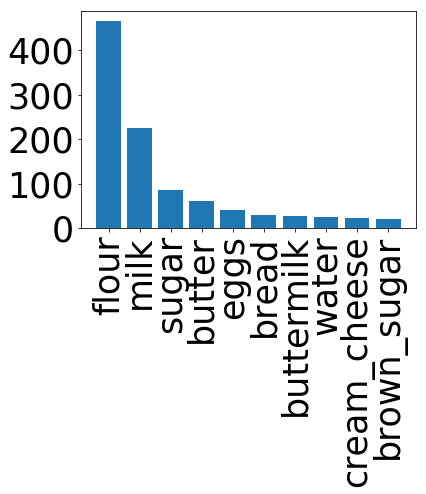}
    \end{minipage}
    \begin{minipage}[t]{.15\linewidth}
    \centering
    \includegraphics[align=t,width=1.0\textwidth]{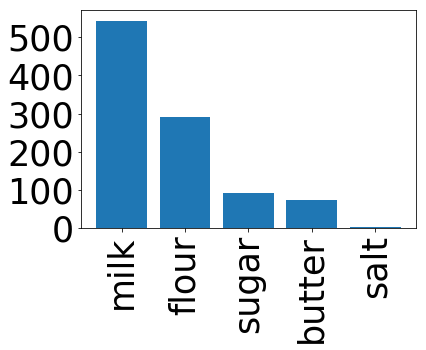}
    \end{minipage}
    \begin{minipage}[t]{.15\linewidth}
    \centering
    \includegraphics[align=t,width=1.0\textwidth]{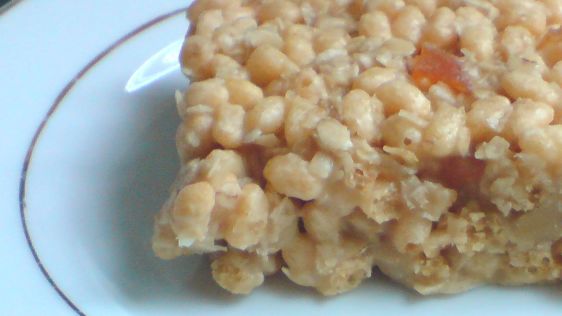}
    \subcaption*{\centering Fruity rice krispie treats/squares-kids no bake}
    \end{minipage}
    \begin{minipage}[t]{.15\linewidth}
    \centering
    \includegraphics[align=t,width=1.0\textwidth]{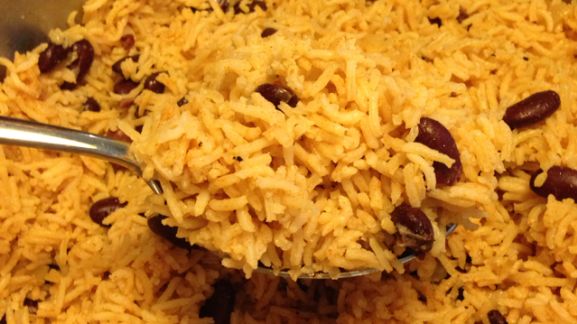}
    \subcaption*{\centering Jagacida(Jag)-beans and rice from Cape Verde}
    \end{minipage}
    \begin{minipage}[t]{.15\linewidth}
    \centering
    \includegraphics[align=t,width=1.0\textwidth]{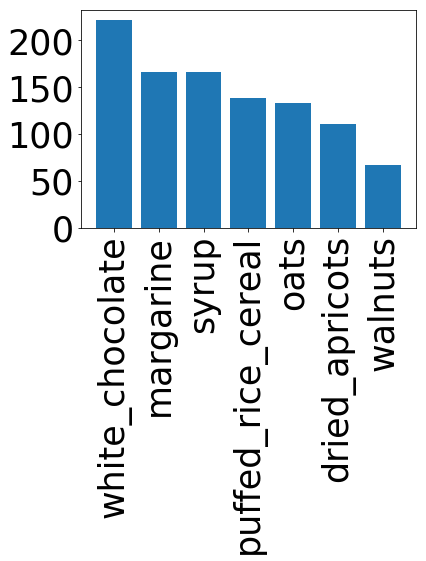}
    \end{minipage}
    \begin{minipage}[t]{.15\linewidth}
    \centering
    \includegraphics[align=t,width=1.0\textwidth]{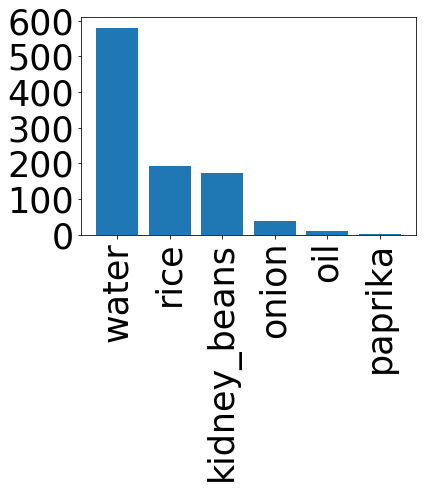}
    \end{minipage}
    \begin{minipage}[t]{.15\linewidth}
    \centering
    \includegraphics[align=t,width=1.0\textwidth]{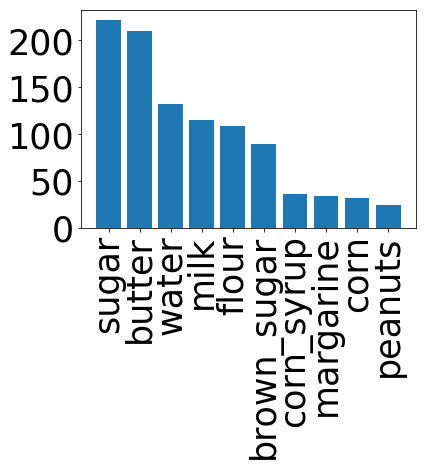}
    \end{minipage}
    \begin{minipage}[t]{.15\linewidth}
    \centering
    \includegraphics[align=t,width=1.0\textwidth]{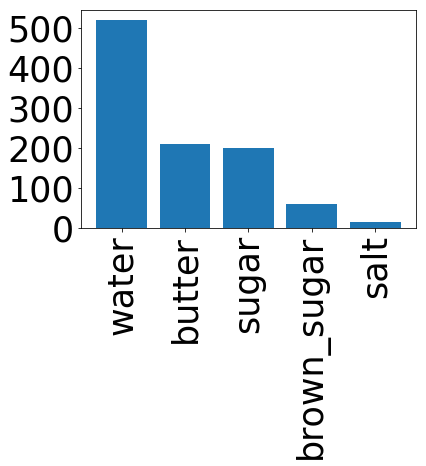}
    \end{minipage}
    \caption{Easy, average and difficult examples. Top row: easy, middle row: average, bottom row: difficult. Columns from left to right: ground truth image, retrieved image, ground truth amounts, retrieved amounts, Softmax results, L1 results. The amounts are normalized to $C=1000$.}
    \label{fig:example}
\end{figure*}

We also demonstrate some easy, average and difficult examples for the models in terms of evaluation metrics. We notice that even the methods perform poorly quantitatively in some difficult test cases, they still produce reasonable ingredient and amount combinations. The result are shown in Table \ref{tab:examples} and Figure \ref{fig:example}.

\begin{table}
    \centering
    \begin{tabular}{|p{1cm}p{1cm}cccc|}
    \hline
         Example&Method&{\bf Recall}&{\bf IOU}&{\bf L1}&{\bf RCE}\\
         \hline
         Easy&Retrieval& {\bf 0.57} &0.33 & 502.74 &{\bf 0.09} \\
         & Softmax & 0.42 & 0.21 & 837.13 & 0.25 \\
         & L1 & {\bf 0.57} & {\bf 0.44} & {\bf 450.37} & 0.16 \\
         \hline
         Average&Retrieval&{\bf 0.38}&{\bf 0.30}&1569.38&{\bf 0.06} \\
         &Softmax&0.25&0.125&{\bf 1537.12}&0.45\\
         &L1&0.25&0.18&1588.95&0.12\\
         \hline
         Difficult&Retrieval&0&0&2000&0.87\\
         &Softmax&{\bf 0.14}&{\bf 0.06}&{\bf 1932.74}&{\bf 0.20}\\
         &L1&0&0&2000&0.45\\
         \hline
    \end{tabular}
    \caption{Quantitative evaluation of easy, average and difficult examples.}
    \label{tab:examples}
\end{table}

In the good and average example, all three methods produce reasonable results. The retrieval system gets foods from the same high-level category: baked goods and prediction-based models capture main ingredients like flour, sugar and butter well. The amount of salt is also accurately predicted. In the bad example, the ground truth uses puffed rice cereals and the similar ingredient rice is also successfully retrieved, although the taste is incorrect: the ground truth is sweet while the retrieved is savory. Softmax successfully predicts the existence of magarine and the overall recipe produced is reasonable as the ground truth image looks like baked goods. L1 successfully predicts a similar ingredient butter with nearly the same amount. Even when the quantitative evaluations are bad, qualitatively the models still give robust predictions.

\section{Conclusion}
We studied a novel problem: given a food image, predict the relative amount of each ingredient needed to prepare the observed food item. Our experiments show that deep model-based prediction methods produce reasonable ingredient and amount predictions; even in the presence of challenging test examples, the methods are still able to yield robust qualitative results.

The problem opens up interesting avenues for future work. First, prediction-based methods leverage amount information during training while the retrieval-based method does not. Nevertheless, there is no significant difference in performance of prediction-based methods and retrieval with ground truth. Further research is needed for improving the performance on this problem. Second, we can leverage amount information in retrieval based methods. Finally, we can adjust the ingredients and the corresponding amounts according to users' dietary needs to generate novel food images.



\appendix
\bibliographystyle{ACM-Reference-Format}
\bibliography{sample-base}

\section{List of Basic Units}
\label{append:unit list}
\begin{itemize}
    \item {\bf Volume Units:} pint, liter, gallon, teaspoon, tablespoon, cup, dash, quart, fluid\_ounce, ml, pinch
    \item {\bf Weight Units:} kg, lb, g, ounce
    \item {\bf Counting Units:} packet, handful, links, sheet, big, package, bunch, clove, leaf, jar, medium, strip, envelope, stick, large, drop, piece, small, container, bottle, head, scoop, of, stalk, glass, sprig, bag, inch, loaf, can, cm, ears, no\_unit, dozen, box, slice, squares
\end{itemize}
"no\_unit" means the ingredient name directly follows the quantity, which is an expression for counting (e.g. 2 apples)

\section{Mapping ingredients to Nutritionix and USDA}
\label{append:mapping}
We pass the list of our ingredients to Nutritionix natural language tagging API\footnote{trackapi.nutritionix.com/v2/natural/tags} and get json formatted responses. The mapping of the ingredient is the "ITEM" field in the response and unit conversion is the "ALT\_MESAURES" field. We then filter out ingredients with no mappings in Nutritionix and ingredients with multiple mappings (e.g."butter flavored shortening" is mapped to both "butter" and "shortening"). For the rest of the results, we first check if the mapping is a sigular or plural form of the ingredient and get a list of ingredients that are not mapped exactly to themselves. For these results, we refine them manually:
\begin{itemize}
    \item {\bf No result:} First we try to query the ingredients in USDA Food database to get the corresponding NDB ID as mapping and the conversion tables. If there is no result in USDA, we look for a synonym for the ingredient and query with Nutritionix to get the mapping and conversion table.
    \item {\bf Multiple mappings:} First we try to select one of the mappings to match the ingredient. If none of the mappings match the ingredient, we query the ingredient in USDA.
    \item {\bf Mismatch:} First we check if the mapping is correct. If it is incorrect (e.g. pea shoots$\rightarrow$pea), we query the ingredient in USDA.
\end{itemize}
We then pass the mappings to Nutritionix nutrients API\footnote{trackapi.nutritionix.com/v2/natural/nutrients} to get the calories per gram with the "serving\_weight\_grams" field and "nf\_calories" field. If the mappings are from USDA, we query the USDA Food database to get calories per gram.

\end{document}